\documentclass[sigconf]{acmart}

\AtBeginDocument{%
  \providecommand\BibTeX{{%
    \normalfont B\kern-0.5em{\scshape i\kern-0.25em b}\kern-0.8em\TeX}}}

\setcopyright{acmcopyright}
\copyrightyear{2018}
\acmYear{2018}
\acmDOI{10.1145/1122445.1122456}

\acmConference{ACM FAT}{2020}{Barcelona, Spain}
\acmBooktitle{FAT* 2020}
\acmPrice{15.00}
\acmISBN{978-1-4503-9999-9/18/06}

\copyrightyear{2020}
\acmYear{2020}
\setcopyright{acmlicensed}
\acmConference[FAT* '20]{Conference on Fairness, Accountability, and Transparency}{January 27--30, 2020}{Barcelona, Spain}
\acmBooktitle{Conference on Fairness, Accountability, and Transparency (FAT* '20), January 27--30, 2020, Barcelona, Spain}
\acmPrice{15.00}
\acmDOI{10.1145/3351095.3372864}
\acmISBN{978-1-4503-6936-7/20/02}

\begin{document}

\title{On the Apparent Conflict Between Individual and Group Fairness}

\author{Reuben Binns}
\affiliation{%
  \institution{University of Oxford, Department of Computer Science}
}
 \email{reuben.binns@cs.ox.ac.uk}

\renewcommand{\shortauthors}{Reuben Binns}

\begin{abstract}
A distinction has been drawn in fair machine learning research between `group' and `individual' fairness measures. Many technical research papers assume that both are important, but conflicting, and propose ways to minimise the trade-offs between these measures. This paper argues that this apparent conflict is based on a misconception. It draws on theoretical discussions from within the fair machine learning research, and from political and legal philosophy, to argue that individual and group fairness are not fundamentally in conflict. First, it outlines accounts of egalitarian fairness which encompass plausible motivations for both group and individual fairness, thereby suggesting that there need be no conflict in principle. Second, it considers the concept of individual justice, from legal philosophy and jurisprudence which seems similar but actually contradicts the notion of individual fairness as proposed in the fair machine learning literature. The conclusion is that the apparent conflict between individual and group fairness is more of an artefact of the blunt application of fairness measures, rather than a matter of conflicting principles. In practice, this conflict may be resolved by a nuanced consideration of the sources of `unfairness' in a particular deployment context, and the carefully justified application of measures to mitigate it.

\end{abstract}



\keywords{fairness, individual fairness, justice, machine learning, discrimination, statistical parity}


\maketitle

\section{Introduction}
Research on `fair' machine learning (hereafter: Fair-ML) has proposed a variety of ways to assess and ensure the fairness of supervised models for prediction and classification. These are largely focused on the application of ML in decision-making contexts which involve allocating more or less positive and negative outcomes to people, on the basis of predictions based on a person's features.

Imagine the following decision-making scenario. An employer must select candidates for interview from a large set of applications. They decide to use a machine learning model which uses information contained in an application and returns a prediction about whether a candidate would make a good employee. Those whose applications have sufficiently positive predictions are invited to interview.

The employer finds that the model is more likely to give positive predictions for one gender (e.g. men) than others (e.g. women and non-binary people). So they decide to apply some Fair-ML technique to the model to prevent this. However, as a result of this modification, a man applicant is not invited to an interview. The applicant complains, pointing to examples of women who were invited to interview despite having qualifications very similar to his. Should the employer continue to interview the women candidates, or adjust its model again to ensure that any `more qualified' men get interviews instead?

Dilemmas like this one have been raised within discussions of Fair-ML to motivate a supposedly important distinction between two competing kinds of fairness. On the one hand, \emph{group} fairness ensures some form of statistical parity (e.g. between positive outcomes, or errors) for members of different protected groups (e.g. gender or race)  \cite{dwork2012fairness}. On the other hand, \emph{individual} fairness ensures that people who are `similar' with respect to the classification task receive similar outcomes.  These measures appear to conflict in cases where, as a result of trying to satisfy group fairness, pairs of individuals who are otherwise similar but differ in a protected characteristic are assigned different outcomes.

The purpose of this paper is to critically assess the apparent conflict between these two kinds of fairness measure. While the debate that follows is primarily theoretical, its implications are significant both for researchers and practitioners:

\begin{itemize}
    \item For Fair-ML researchers, the apparent conflict between individual and group measures continues to motivate new research papers many years after it was initially formulated (in \cite{dwork2012fairness}). While many such papers propose technical means to reconcile the two measures, they generally lack sustained theoretical discussion regarding the underlying principles supposedly in conflict.
    \item For practitioners, as ML models continue to be applied in a wide variety of applications in the public and private sector,  decision-makers will turn towards governance measures proposed by the Fair-ML community. In particular, the intuitions which motivated the proposal of individual fairness in the academic debate - namely, that `less qualified' individuals may unfairly be given opportunities in order that the model can acheive some statistical measure of group fairness - may lead some to object to particular implementations of group fairness in practice. It is therefore essential that organisations deploying such measures can provide sound justifications for the selection of particular fairness measures.
\end{itemize}

Such justifications will require careful consideration of the principles behind, and relationship between, group and individual fairness. The contention of this paper is that the supposed conflict between the measures does not arise at the level of principle; these are not two fundamentally different and conflicting types of fairness. Rather, the appearance of conflict between the two is an artefact of the failure to fully articulate assumptions behind them, and the reasons for applying them in a particular context.

\subsection{Overview of argument and contributions}
Section 2 introduces the notions of individual and group fairness as they have been developed in the Fair-ML literature, and elucidates their perceived merits and shortcomings, and explains why it has generally been assumed that they are in tension with each other. It also provides an overview of various recent proposals to minimise the trade-offs between the two measures.

Section 3 aims to map the relationship between these fairness measures and a range of concepts and theories from moral, political, and legal philosophy that they might be plausibly thought to correspond to. I argue that while there are differences in the way individual and group fairness are applied in specific contexts, they don't necessarily correspond to distinct and conflicting \emph{principles}. I argue that, at this abstract level, individual and group fairness are not only \emph{not} in conflict, but are in fact just different ways of reflecting the same set of moral and political concerns.

In section 4, I pursue a complementary line of argument, which examines a possible underlying motivation for individual fairness, namely: \emph{individualized} or \emph{particularized} justice. This is the idea that people deserve to be treated as unique individuals, and assessed on a case-by-case basis. Like individual fairness, this concept is grounded in the treatment of individuals. However, on this view, even supposedly individually-fair models can be seen as individually unjust because they still generalise between people who share the same features. In this sense, what passes for individual fairness actually amounts to a special case of group fairness. So again the apparent distinction disappears when considered at the level of principle.

Section 5 aims to persuade the reader that in so far as fairness measures can be applied in particular contexts, there are very few such contexts in which what have been termed `individual' and `group' fairness measures would be simultaneously applicable and conflicting. In so far as they may appear to conflict in a given circumstance, this will be down to unstated conflicting moral and empirical assumptions regarding the purpose of the decision-making procedure and the nature of the fairness concerns inherent in that particular context. This section also connects this argument to a similar one made in \cite{friedler2016possibility}.

In the interests of promoting interdisciplinary dialogue, to satisfy readers from both computer science and social science / humanities, I avoid formal notation, in favour of prose descriptions of technical definitions.


\section{`individual' and `group' fairness in Fair-ML}

This section introduces the notions of individual and group fairness as they have been developed in the Fair-ML literature, and elucidates their perceived merits and shortcomings, and why it has generally been assumed that they are in tension with each other. It also provides an overview of a variety of recent proposals to minimise the trade-offs between the two measures.

\subsection{Initial formulations of group and individual fairness}

Research on bias, fairness and discrimination in socio-technical systems has a long history which significantly predates specific work on fairness in machine learning (see e.g. \cite{gandy1995s,friedman1996bias}). However, such work did not propose formalised quantitative measures of fairness.\footnote{Formalised definitions of fairness which foreshadow those in Fair-ML did appear in areas other than ML, such as testing in education and hiring. See Hutchinson \& Mitchell 2019 \cite{hutchinson201950}.} The first examples of Fair-ML fairness definitions arose in the field initially known as `discrimination-aware data mining' in the mid 2000's.

\subsubsection{Group fairness measures}

These early papers used fairness measures based on statistical parity between protected groups (e.g. gender, race) in each outcome class, and hence are classed as group fairness measures \cite{pedreshi2008discrimination,calders2009building}.

According to statistical parity, a classifier is fair if there are equal proportions of each protected group in each outcome class. A wider range of different group fairness measures (also known as `statistical' measures) have since been proposed. They are all based on some measure of statistical parity between people who have different values for a set of protected attributes. Some look at the distribution of errors between groups, e.g. false negative and positive rates \cite{chouldechova2017fair}. Other group fairness measures focus on calibration: e.g., a model is fair if it is equally calibrated between members of different protected groups (where calibration measures how closely the model's estimation of the likelihood of something happening corresponds to the actual frequency of the event happening) \cite{kleinberg2016inherent}.


One problem that is often raised for group fairness measures is that they are only suited to a limited number of coarse-grained, prescribed protected groups. They may miss unfairness against people as a result of their being at the intersection of multiple kinds of discrimination \cite{crenshaw1990mapping}, or groups which are not (yet) defined in anti-discrimination law but may need protecting \cite{wachter2019right}. If group measures are only applied to coarse-grained groups separately (e.g. gender, race), they might permit unfairness for structured combinations of those groups (e.g. black women), also known as \emph{fairness gerrymandering} \cite{kearns2017preventing}. Some proposals aim to solve this problem by defining fairness across many different combinations of protected characteristics. A challenge here is that testing on every possible subgroup doesn't scale well to the large number of possible combinations, and may lead to overfitting \cite{kearns2017preventing,hebert2017calibration}.  While these works are in part motivated by notions of individual fairness, since they still deal with groups (albeit potentially very fine-grained intersectional groups) they can still be seen as a form of group fairness measure. 

\subsubsection{Individual fairness measures}

The first recognisable version of individual fairness as a kind of Fair-ML criteria was proposed in \cite{dwork2012fairness}. The motivation for individual fairness in this paper was the concern that simple statistical parity between protected groups in each outcome class could be intuitively unfair at the individual level. The authors present several examples in which `statistical parity is maintained, but from the point of view of an individual, the outcome is blatantly unfair'. They note that a classifier could be fair if it just gives positive outcomes to candidates from certain protected groups at random, just to `make up the numbers'. This would be unfair to those individuals who fail to get a positive outcome which they deserve because they are more `qualified' than the randomly selected members of the otherwise under-represented group (these may include both members of the advantaged and the disadvantaged groups). This is also to the detriment of the decision-maker who loses utility, in the sense that a fair classifier on this definition needn't provide \emph{accurate} predictions or classifications.

To avoid these problems, various papers have proposed an individual-level fairness measure \cite{dwork2012fairness,joseph2016fairness, lahoti2019ifair}. The intuition behind this measure is that people who are similar with respect to the task should be given similar predictions or decisions. Individuals are defined in terms of a distance metric which represents how similar they are to each other with respect to the features related to the task or context of decision making. Two individuals are alike if their combinations of task-relevant attributes are `nearby' each other in the defined metric space. It is assumed that the distance metric is somehow defined by the people who set the policy (e.g. college admissions tutors), and possibly with broader societal agreement, and that this can be applied to individuals (e.g. college applicants). Individual fairness says that for any two individuals, if their distance in task-relevant similarity is sufficiently small, they should receive the same outcome (as defined in \cite{dwork2012fairness}); or alternatively, that  `less qualified individuals should not be favored over more qualified individuals' \cite{joseph2016fairness}.

While a mapping of the task-relevant similarity between all individuals could potentially avoid the problems with statistical parity, it is unclear how decision-makers are supposed to obtain it. This is the major shortcoming associated with the initial formulation of individual fairness, one acknowledged by the authors \cite{dwork2012fairness}. Even if such a mapping could be obtained, if it is only defined over individuals in the training set, we also lack a way to generalise to new, unseen, individuals.

This difficulty has arguably stymied the practical application of individual fairness measures, despite their theoretical appeal. However, some subsequent proposals aim to relax the need for a policymaker to provide all distance mappings between individuals, and instead provide methods to \emph{learn} such a distance function. In some cases, the distance function is learned directly from the data (e.g. \cite{zemel2013learning,lahoti2019ifair}). In Zemel et al. 2013, the approach involves mapping individuals to a set of clusters based on their features, and these cluster-based `prototypes' are used as a relaxed proxy for individual similarity \cite{zemel2013learning}. In this way, new examples could be mapped automatically without requiring \emph{a priori} similarity metrics to be defined by a policymaker. Similarly, Lahoti et al. 2019 propose a technique based on clustering, which aims at minimising the distances between the non-protected attributes of pairs of individuals receiving the same outcome \cite{lahoti2019ifair}. In other cases, the distance function is learned from judgements elicited from a panel of ethical experts in response to comparisons between pairs of candidates \cite{jung2019eliciting}.

Other approaches give up the attempt to determine a similarity metric, and focus on the notion of individual merit \cite{joseph2016fairness}. A model is fair on this view if the probability of an individual being predicted to have a certain label corresponds to the true (but unknown) probability that they actually have (or will have) that label. For instance, a CV scoring model which predicts a job applicant has a certain chance of succeeding at interview should reflect the true probability the job applicant will actually succeed. The challenge for this approach is that, since the true labels of new instances are unknown, strong assumptions have to be made about the relationship between the observable features and the unobservable ground truth.

\subsubsection{Reconciliation between individual and group measures}


Several proposals attempt to find an optimal balance between the two kinds of measures. For instance, in their proposal for combining individual and group fairness, Zemel et al. build in statistical parity constraints into the process of learning fair representations, such that an individual's chance of being represented as a particular prototype does not depend on their membership in a protected group \cite{zemel2013learning}. While Lahoti et al's explicit aim is to capture individual fairness, they note that their approach indirectly improves group fairness because it reduces information on protected attributes, and, they argue, makes the multi-objective problem of satisfying utility, group fairness and individual fairness more tractable \cite{lahoti2019ifair}.

However, despite attempts to partly reconcile these two measures, it appears that there are inherent trade-offs between them. Unless the distribution of features and labels is the same across different protected groups, the intuition behind individual fairness will always appear to be violated in some way. According to proponents of individual fairness, we should remain concerned about models which give different outcomes to people who are similar, whether their similarity is defined according to some distance metric, or in terms of the true probability of having a certain label. Lahoti et al regret that individual fairness has been generally neglected in favour of group fairness, because the former is `intuitive and captures aspects that group fairness does not handle' \cite{lahoti2019ifair}. Conversely, for proponents of group fairness, individual fairness is liable to result in seemingly unjust disparities in outcomes between protected groups.

\section{Is there a group / individual distinction in principle?}

This section aims to connect the justifications offered for individual and group fairness measures in the Fair-ML literature to a range of concepts and theories from moral, political, and legal philosophy that they might correspond to. Based on this mapping exercise, I argue that while individual and group fairness may appear to conflict in the Fair-ML literature, they don't necessarily correspond to distinct and conflicting philosophical or moral principles. I argue that, at this abstract level, individual and group fairness are not only non-conflicting, but are in fact different aspects of the same consistent set of fundamental moral and political concerns.

Before proceeding, it is worth acknowledging that it may seem surprising that an appeal to philosophy could \emph{reduce}, rather than increase, the complexity and nuance of normative concepts formalised by computer scientists. From the perspective of computer scientists and other technical researchers working on Fair-ML, it may appear that philosophers, lawyers, and other humanities scholars deal with many different and conflicting notions of fairness that have been the subject of intractable debate for millenia. Given this, the variety of fairness measures proposed by Fair-ML researchers may seem positively parsimonious. Similarly, critics of Fair-ML from philosophical and other backgrounds might reasonably worry that in technical researchers' haste to formalise, the nuances and differences between moral concepts might be flattened into an artificially small cluster of homogeneous technical definitions of fairness. As such, one might expect that social science and humanities scholars are more likely to expand sub-types of fairness rather than collapse them.



It may be true that there are fewer normative concepts dealt with in Fair-ML, than dreamt of in philosophy. But in some cases, philosophical and social science perspectives may illustrate that concepts which are apparently conflicting at a technical level are not reflective of a deeper normative conflict. This paper argues that this is the case regarding the notions of individual and group fairness as they have developed in the Fair-ML literature. That is, while these measures are presented as representing two different kinds of fairness, they can just as easily correspond to the same underlying moral concept. Any conflicts between the two would therefore be artefacts of the possibility that the same underlying concept can be operationalised in different ways.

In this section I briefly outline two norms which might plausibly be thought to lie behind these notions. In the interests of brevity, the selection and presentation of these norms is necessarily partial, and draws from a narrow literature. However, in searching for corresponding concepts in the literature, I take as my starting point the partially articulated normative motivations mentioned in the Fair-ML literature from which the individual / group distinction arose. While these papers did not generally engage in sustained consideration of corresponding and supposedly conflicting norms behind individual / group fairness, they do provide brief and sometimes oblique references to normative concepts.\footnote{ More recent work draws more explicit connections to political philosophy e.g. \cite{binns2017fairness,hoffmann2019fairness,heidari2019moral}} 

\subsection{Consistency}
In Dwork et. al 2012 \cite{dwork2012fairness}, individual fairness is motivated by reference to the notion that `similar cases should be treated alike'. This is a position associated with Aristotle's conception of justice as consistency \cite{schauer2018treating}. It is valuable desiderata of justice that decision-makers can produce a single predictable and correct judgement in each case; when presented with identical cases, judges ought usually to come to the same answer. For Dicey, this was to be acheived by deterministic application of the rules \cite{dicey2013law}; for Dworkin, judges have more flexibility in their interpretation of rules and principles, but nonetheless ought to alight on a single judgement \cite{dworkin1978no}. The implication is that given two cases in which the features are identical, and assuming an identical legal system with the same set of statutes, case law and institutional history, judges ought to come to the same conclusion.\footnote{Not all of jurisprudence places such a high premium on consistency; there is a large literature on the desirability of discretion, and scepticism about Dworkin's faith in there always being a right answer. See the section below on individual justice.}

From one perspective, consistency should not be considered a problem for ML-assisted decision making. Supervised machine learning models are generally speaking deterministic in producing identical outputs in response to particular inputs (at least, until they are replaced by new iterations trained on fresh data). So on one level, both group-fair and individual-fair models satisfy consistency. However, Aristotle's maxim bears a striking similarity to the initial formulation of individual fairness in Dwork et al, in terms of `similar individuals being treated similarly'. If we apply Aristotle's maxim to comparisons between pairs of people who have been classified differently due to a group fairness constraint, despite having similar features, it may seem that some version of the consistency principle has been broken. In this context, \emph{consistency} appears to be the norm to which individual fairness corresponds.

The problem with this formulation of consistency is that, in not specifying the means of comparing likeness between cases, it tells us very little and leaves everything to be debated. This has lead some to criticise Aristotle's notion of equality as being `empty' \cite{schauer2018treating, westen1982empty}. Two individuals from different protected groups can only be counted as `similar' if we omit their protected characteristics; but what justifies this omission? Why focus on some similarities and not others? What is the justification for designating certain characteristics as protected in the first place? The mere principle of consistency cannot answer these questions; both individual and group fairness could therefore be seen as `inconsistent' depending on which features we exclude or include. However, these questions are dealt with in accounts of egalitarian political justice, covered in the next section.

\subsection{Egalitarianism}

Alongside the indirect references to Aristotle's notion of `treating like cases alike', a handful of early Fair-ML papers (inlcuding \cite{dwork2012fairness}) refer to political philosophers such as John Rawls, who propound theories of egalitarianism. Egalitarianism is the notion that people need to be treated in some senses equally, or that certain things need to be distributed according to some egalitarian principle \cite{cohen1989currency,rawls2009theory,dworkin1981equality,sen1992inequality,anderson1999point,walzer2008spheres}. An overview of egalitarian theories of justice and their relationship to Fair-ML is beyond the remit of this paper (for which, see \cite{binns2017fairness}). While egalitarians advocate for equality, they usually do not mean that everyone should be allocated an equal share of everything regardless of their choices and circumstances. Rather, they aim to ensure a `level playing field' for everyone (often referred to as `equality of opportunity'). A prominent approach, known as luck egalitarianism, holds that inequalities between people are only just provided they are not the result of brute luck \cite{huseby2016can,temkin1986inequality,arneson1989equality}. If people find themselves worse off through no fault of their own, luck egalitarianism would consider their plight to be unjust. If on the other hand, people are worse off as the result of free choices the consequences of which are reasonably foreseeable, or if they take an informed risk which doesn't pay off, this is not an injustice on the luck egalitarian account. 

Egalitarians may differ in their views about which attributes or circumstances people can rightly be held responsible for. They may disagree, for instance, about whether the distribution of natural talents should be regarded as a matter of luck and therefore unjust \cite{roemer1985equality}, or instead whether people should be permitted to reap the rewards of their natural talents \cite{rawls2009theory}. Some critics question the notion of a `luck'/'choice' distinction, pointing out that certain disadvantages may be the result of free and informed choices, but for which the chooser should not be left significantly worse off as a result, such as choosing to forgo a more lucrative career in order to help others \cite{anderson1999point,thaysen2017bad}. Despite these differences, the basic impetus remains the same; people should only suffer disadvantages, or enjoy advantages, that result from certain kinds of choices they make, and not circumstances they find themselves in through no fault of their own.


When it comes to the allocation of jobs, welfare, loans, places in higher education, or bail decisions, egalitarian theories could provide a philosophical framework within which to evaluate the justice of decisions. For instance, we might use a luck egalitarian account to assess the extent to which decision outcomes can vary depending on certain features. As proposed in recent work by Heidari et. al \cite{heidari2019moral}, features can be separated into those for which individuals are not responsible, and those for which they are. (The authors use the terms `circumstance' and `effort' to demarcate these, but depending on the variant of egalitarianism adopted, different terms may be appropriate.) For instance, to the extent that certain attributes held by college applicants are the result of brute luck (e.g. whether or not they were born to parents could afford to pay for expensive test tuition), such attributes would not justify more favourable outcomes in admissions decisions. Such a direct application of philosophical accounts of justice to Fair-ML might not always be possible; for instance, those cited above who reject liberal egalitarian tenets like the choice / circumstance distinction \cite{young2010responsibility,hooks2014ain,cooper2016intersectionality} might be sceptical about the possibility of ever finding features which are untainted by oppressive structures.

However, to the extent that such egalitarian theories \emph{can} meaningfully be a guide to `fair' feature selection, they would typically classify the kinds of protected characteristics commonly cited in Fair-ML - such as gender or race - as attributes which would \emph{not} justify unequal distributions. In other words, concerns about unfairness on these specific grounds are a subset of a broader set of egalitarian concerns (albeit a highly significant, and perhaps paradigmatic subset). In so far as egalitarianism seeks to equalise the distribution of outcomes between such groups, it might seem to be the natural corresponding normative principle that would justify group fairness.

However, just as the principle of consistency can actually be commensurate with group fairness (despite its natural affinity with individual fairness), so egalitarian concerns can also be couched in terms of individual fairness (despite their natural affinity with group fairness). Consider that when specifying an individual fairness metric, the policymaker will need to consider certain features (e.g. exam scores) and ignore others (e.g. first language) when assigning distances between pairs of individuals. Those choices reflect normative assumptions which may well correspond to the egalitarian principles above. This make sense if we see the notion of `task-relevance' as \emph{already} incorporating normative aspects. For instance, a policymaker operationalising the luck egalitarian framework could base their determination of task-relevant similarity on norms and causal assumptions which account for the attributes for which individuals can be held responsible for. In some cases, they might adjust the distance metric to account for features which reflect circumstances outside the individual's control which would otherwise make them appear `less qualified'. For instance, they might determine that while person A has the same exam score as person B, since A took the exam in their second language, A may be more qualified for the task than B (because A's exam score suggests they must have higher overall competence, and/or have applied greater effort, given their language barrier). Such an approach to individual fairness would incorporate the kinds of structural unfairness that egalitarians might be concerned about, while still operating at the individual level.

The considerations above suggests that while there may be natural affinities between the usual formulations of individual fairness and consistency, and the usual motivations for group fairness and egalitarianism, these are only surface deep. Consistency and egalitarianism themselves do not conflict at the level of principle. In fact they can even be seen as mutually implied; in so far as luck egalitarianism aims to remove luck from allocation, it implies consistency. And in so far as the process of defining task-relevant similarity can already be an exercise in normative judgement, it makes sense to incorporate egalitarian concerns into it. It is therefore both possible and indeed coherent to adopt consistency-respecting formulations of group fairness, as well as egalitarian formulations of individual fairness. The details of their implementation may vary, but both approaches can reflect the same set of normative motivations.

\section{Individual Fairness vs Individual Justice}

Those with a strong commitment to the notion of individual fairness as distinct from group fairness may find this attempt at reconciling the two measures unsatisfying. They might appeal to the idea that there is a fundamental normative difference between a decision-making procedure which treats people differently on the basis of group membership, and one which focuses on them as \emph{individuals}.

What principle might lie behind such an appeal to focus on individuals, not groups? In this section, I consider a more fundamental notion that could plausibly be thought to motivate individual fairness. This is what might be called `individual justice'; the idea that individuals should be assessed on their \emph{own} qualities, circumstances, and attributes, not on the basis of generalisations about groups of which they happen to be a member. However, despite the apparent connection, proponents of individual fairness will not find solace in this notion, for reasons covered in the following section. Individual fairness measures ultimately, like group fairness measures, fail to treat people truly as individuals.

\subsection{Aristotle's other maxim}

To understand this notion of individual justice, we can appeal to another one of Aristotle's maxims. In addition to saying that like cases should be treated alike, Aristotle described a different (and possibly contrary) position (as quoted in \cite{schauer2018treating}):

\begin{quote}
`There are some things about which it is not possible to pronounce rightly in general terms ... the raw material of human behaviour is of this kind'.  
\end{quote}

Legal philosopher Frederick Schauer calls this notion `individualised' or `particularised' justice \cite{schauer2018treating}; it holds that people need to be assessed individually, rather than on the basis of rules derived from consideration of similar cases who came before. In German jurisprudence, it is known as \emph{Einzelfallgerechtigkeit} (`justice in the particular case') \cite{britz2008einzelfallgerechtigkeit}. Similarly, U.S. jurors have argued that making a decision about one individual on the basis of a generalisation about similar people `is not consistent with respect based on the unique personality each of us possesses'.\footnote{In Rice v. Cayetano , 528 U.S. 495, 517 (2000). See also \cite{armour1994race,walen2010unified} --- I thank Ren\`ee Jorgensen Bolinger for bringing these works to my attention via forthcoming work \cite{bolinger2019what}.} A similar notion is deployed in some philosophical accounts of discrimination, according to which it is wrong because it fails to treat people as individuals \cite{beeghly2018failing,lippert2014born,moss2018probabilistic}.

Arguments for the normative value of individualised assessment are not the preserve of jurisprudence and philosophy. They often appear in calls for public administrators to exercise discretion rather than routine application of rules \cite{pratt2009brief}, and also in many contexts in which issues of fair treatment arise between private actors (e.g. housing). In contexts in which public administration and private decision-making are being automated or augmented with algorithms, the need for occasional individualised consideration is one of the reasons for keeping a human-in-the-loop, or `screen-level bureaucrat' on hand \cite{citron2007technological,bovens2002street,veale2018fairness,alkhatib2019street}. By scrutinising additional information about the individual that the algorithm does not consider, and considering alternative forms of reasoning regarding mitigating circumstances that an algorithm could not, the `human-in-the-loop' may be able to serve the aim of individual justice.\footnote{For a sceptical appraisal of such claims, see \cite{binns2019human}.}

\subsection{Individual justice and ML}

The focus on individual cases within individual justice might appear to provide proponents of individual fairness with a new argument for favouring their approach over that of group fairness. It suggests that any form of generalisation on the basis of group membership is unfair. If we really care about being fair to individuals, then we should look to \emph{individualized} justice.

However, individually-fair models would also count as individually unjust on this account. In its algorithmic guise, individual fairness still essentially equates an individual with their features (or, perhaps, their position in the model-defined feature space). Once a similarity metric has been defined (or whatever other individual fairness approach chosen has been applied), features selected, and an ML model trained and deployed, any individuals who share the same features will get the same outcome. They are given a prediction or classification on the basis that they are like those who came before them who shared the same features. According to individual justice, this is unfair; individual cases must be assessed on their own, disregarding any previous knowledge that may have been inferred from previous similar cases, and potentially incorporating new kinds of information and reasoning particular to the case.

This suggests that if we really care about individualized assessment in a decision-making context, then we cannot treat it principally as a `decision problem' in the sense that computer scientists might think about decision problems; namely, as yes-or-no questions that can be provided in response to a delimited range of possible input values. While deeply impractical for anyone attempting to replace human decision-making with algorithms, individual justice is a logical extension of the argument that generalisation based on group membership is wrong. In so far as individual fairness is at all viable as a kind of Fair-ML measure, it cannot be truly individualized, and is in fact akin to a kind of group generalisation, where each group is defined in terms of the people who occupy a particular point in the defined task-relevant metric space.

To recap the argument of this section: proponents of individual fairness might think that by drawing on the notion of individual \emph{justice}, they have a more fundamental argument in favour of their own position and against group fairness. However, even individually-fair models would still involve generalisation (albeit more fine-grained) and therefore do not really preserve individual justice.

\begin{table}
  \label{tab:freq}

\begin{tabular}{c|c}
    \toprule
    \emph{\textbf{Fair-ML concept}} & \emph{\textbf{Philosophical Corollary}}\\
    \midrule
    Group Fairness & Egalitarianism, Anti-Discrimination\\
    \midrule
    Individual Fairness & Consistency, Individual Justice*\\
    
  \bottomrule
\end{tabular}
\caption{Putative relation between Fair-ML concepts and legal / philosophical corollaries as intimated in existing Fair-ML literature. However, as illustrated in Table 2, both families of fairness measure can equally be related to each of the other philosophical corollaries. * Individual fairness only superficially reflects individual justice - see section 4.2 }
\end{table}

\section{Dissolving the conflict in practice}

The previous sections argued that there is no fundamental conflict between two of the main principles that plausibly correspond to individual and group fairness measures (namely, consistency and egalitarianism), and furthermore, that neither individual nor group fairness can respect the essence of a third principle (individual justice), despite its surface similarity with the former. Having dissolved the group / individual conflict in theory, this section aims to illustrate how it can be dissolved in practice, through two specific hypothetical examples.

Those familiar with trying to implement Fair-ML techniques may suspect that, despite the theoretical compatibility argued for above, we would still face difficult normative conflicts between these two types of fairness measures when applying them in practice. The examples presented below aim to show that such conflicts are not primarily the result of selecting individual or group fairness measures. Instead, they are likely to be the result of unstated conflicting moral and empirical assumptions regarding the decision-making context. Once these conflicting assumptions are resolved, the resulting fairness concerns can be reflected in \emph{either} individual \emph{or} group measures of fairness --- but there will usually be no need to pit one measure against the other.



\subsection{Example 1: College Admissions} ~\\
Imagine the decision-making context of \emph{college admissions}, where admissions panels must decide which applicants to accept for a degree programme.\footnote{The use of this example should not be taken to condone the use of algorithmic decision-making to make such important decisions; it is for illustrative purposes only, as in other works, e.g. \cite{friedler2016possibility}. We also note that, unfortunately, such systems are already in use \cite{ekowo2016promise}.} This is a context in which individual and group fairness measures might be thought to be in conflict.

If a higher proportion of applicants from group B are accepted than applicants from group A, the system might seem unfairly biased against group B (where A and B are different values of a protected characteristic). On the other hand, if two applicants who have similar SAT scores, extra-curricular activities, and interview scores are given different outcomes, the system may seem unfair in a different way. An unsuccessful applicant from group B who has equal or better scores than some successful members of group A might feel they have been dealt an unfair outcome (for reasons expressed in Aristotle's `like cases' maxim).

A typical Fair-ML approach might be for the college admissions panel to address the first set of concerns through a group fairness metric, and the latter through an individual fairness metric. They would then be faced with a trade off between these two, to be dealt with by picking one or the other, or attempting to strike a balance by treating the learning process as a multi-objective optimisation problem. However, the arguments presented above suggest that the two sets of concerns may actually have a common normative source; and that any conflict between the measures is more likely to be the result of a failure to clarify what kinds of injustice are assumed to exist and are being addressed.

The first task is to decide what objectives they have in awarding places to applicants. One criteria might be giving places to applicants most likely to succeed in their college studies. This is a laudable aim, but in most cases it is not and should not be the only aim. They might also aim (echoing the luck egalitarian perspective), to avoid making decisions which would lead people to suffer disadvantages arising from circumstances outside of their control. Some applicants might appear more or less likely to succeed at college due to factors like whether or not they had parents who could pay for expensive test preparation. They may also actually be more or less likely to succeed after being admitted due to factors outside their control, such as systemic racism or sexism in the higher education institution. In such cases, it would be unjust to distribute college places on the basis of features which predict success at college, in so far as such features and such success are in part a function of circumstances that should be excluded (as would be suggested by the egalitarian perspective).

If the college admissions panel can agree upon a set of assumptions about what factors lie outside an applicant's control, and the probable influence of those factors on the distribution of features in college applications, they might then be in a position to formulate some kind of fairness measure to apply to their system. Of course, this is not an easy task, and the legitimacy of any assumptions made could (rightly) be challenged. However, assuming for the sake of argument that the college has a legitimate and accountable process by which they can reach an agreeable set of assumptions, the first question they should ask is not: `should they use individual or group fairness?'. Rather, it is: `what kinds of injustice do we believe may be in operation in this context that may be reflected in and perpetuated by the model being used?' Two ends of the spectrum of answers to this question are a) assuming the disparities are mostly due to structural discrimination, and b) assuming they are mostly due to individual choice, not circumstance. Let us consider each in turn and the consequences for the choice of fairness measure. \newline

\emph{Assuming SAT and graduation rate disparities are due to structural discrimination}

Imagine they conclude that applicants from group A on average face greater adversity in society in general, as well as more specifically in the context of college admissions and post-admission performance. In order for applicants from group A to obtain the same SAT scores as applicants from group B, they have to make greater sacrifices and apply more effort. When admitted, they face more barriers to success than group B, which affects their chances of graduation. While they cannot be sure exactly how great the effect these forms of adversity is, the college admissions panel comes to the conclusion that, for the purposes of implementing Fair-ML, the effect is largely responsible for any observed differences in SAT scores and graduation rates between groups A and B.

Given this conclusion, they could equally choose a group fairness \emph{or} individual fairness measure. If we assume that differences in SATs and graduation rates are largely responsible for the disparate impact of the original ML model, then a statistical parity measure might be appropriate. Similar assumptions could justify other kinds of group fairness measures, such as equalised false positive / false negative rates. 

But the same assumptions could equally motivate the adoption of an individual fairness measure which factors in those forms of discrimination. For instance, if they have agreed that the adversity faced by applicants for group A accounts for an average of 100 fewer points on their SATs, they can factor this in when determining if pairs of individuals from groups A and B are similar. An applicant from the A-group with an SAT score of 1500 will be considered to be similar to an applicant from the B-group with an SAT of 1600.

Indeed, this idea is posited in the original formulation of individual fairness \cite{dwork2012fairness}, where the authors note that individual fairness `will most likely only be society's current best approximation to the truth'. They `envision classification situations in which it is desirable to ``adjust'' or otherwise ``make up'' a metric', citing the example of college admissions offices `adding a certain number of points to SAT scores of students in disadvantaged groups'. Similar adjustments could be made to other features used for prediction. So just as with group-based statistical parity measures, the individual fairness distance metric could represent a rough consensus about how disadvantaged certain groups are in a given context. In some contexts, it may be that the simplest and most workable assumption is that the effects of disadvantage faced by a given group fully account for any negative disparities reflected in the data for that group (therefore justifying a statistical parity measure). \newline
\\

\emph{Assuming SAT and graduation rate disparities are due to personal choice}

On the other hand, imagine instead that the college admissions board comes to a rough consensus that the differences between groups A and B are due to the personal choices of members of each group and not due to circumstance outside their control. They conclude that individuals from group B face no less adversity than those from group A, and the differences between outcomes are accounted for by other factors (perhaps members of group B just spend more time on their applications on average, and study harder after admission).

Again, given this assumption, the choice is not between individual or group fairness measures, but rather between a range of measures which reflect these assumptions, some of which are classed as `individual' and others which are classed as `group'. For instance, they could choose an individual fairness measure which doesn't make adjustments to SAT scores on the basis of membership in group A. Or they could choose a group fairness measure which accepts that differential base rates between A and B are not problematic, and not attempt to correct for these. For instance, they could choose `equal calibration', according to which the model should be equally well-calibrated on applicants from groups A and B \cite{kleinberg2016inherent}. Equal calibration allows for models which give disproportionately positive / negative predictions between groups, so long as those proportions are reflected in the real labels of groups A and B in the test data set. In other words, it treats disparities in SATs or graduation rates, and any corresponding disparities in predictions by a model trained on them, as fair. On this approach (assuming that higher SAT scores have a positive monotonic relationship with graduation rates), then an applicant from group B would likely not (other things being equal) lose `their' place to a `less-qualified' applicant from group A. This therefore echoes the original motivation for (un-adjusted) individual fairness.

\begin{table*}

\begin{tabular}{c|c|c}
    \toprule
    \emph{\textbf{Normative/Empirical Assumption}} & \emph{\textbf{Individual Fairness}}& \emph{\textbf{Group Fairness}}\\
    \midrule
    Disparities due to personal choices & `Raw' Similarity Metric & Equal Calibration\\
    \midrule
    Disparities due to unjust structures & Group-adjusted Similarity Metric & Statistical Parity\\
    
  \bottomrule
\end{tabular}
\caption{Possible normative/empirical assumptions, and corresponding group/individual fairness measures.}
\end{table*}

\subsection{Example 2: Financial Lending}


Imagine a decision-making scenario in which a bank uses an ML model to determine which loan applications will be approved or denied. Imagine the bank notices that historically, men have had a higher probability of being given a positive label (indicating that they repay their loan) than women.\footnote{This section has been informed by analysis of the Statlog (German Credit Data) Data Set \cite{Dua:2019}. Comparing the positive labels of women and men in this dataset shows a statistical (dis)parity of -11\%}\newline

\emph{Assuming positive label disparities are due to historic discrimination}~\newline 
After consulting with a range of stakeholders and experts, the bank identifies several likely reasons why women in the dataset have historically been less likely to receive a positive label. These include factors like (for instance) clerks being more lenient with repayment deadlines for men, and women being more likely to be single parents with unpredictable outgoings due to their dependants. The bank comes to the conclusion that it would be unfair for these factors to disadvantage future women applicants. On the basis of the evidence it has been able to gather, it assumes that the differences in outcomes between men and women, and the features which predict them, can be accounted for by these unjust structural disparities.

Given this assumption, should the bank use group or individual fairness measures to evaluate the fairness of its lending algorithm? If it opts for group fairness, this could involve, for example, re-weighting the historical data to edit feature values to increase group fairness [41]. Alternatively, the bank could opt for individual fairness. This would involve creating a task-relevant similarity metric and applying it to the data using one of the methods described in the individual-fairness literature (e.g. \cite{dwork2012fairness,jung2019eliciting,lahoti2019ifair}). Having already concluded that the features that predict higher default rates by women (namely, `late' repayments and less predictable outgoings) are features which are unjustly affected by gendered social structures, the values of these features for women will need to be adjusted accordingly when applying the individual fairness metric (as with the SAT score adjustment example from \cite{dwork2012fairness}, and above).

While these individual and group approaches may involve somewhat different processes, they are equally motivated by the same set of empirical and moral assumptions - i.e. that the disparity in default rates between men and women reflect unjust structures - and both aim to produce a model which makes predictions which ignore the effects of those social structures. Both approaches could result in men objecting that they have lost a loan opportunity to a `less qualified' woman - so the supposed advantage of individual fairness (that it avoids such individual scenarios) is lost once the structural dimension is factored into an individual fairness measure.\newline 

\emph{Assuming positive label disparities are due to personal choices, not circumstance}\newline 
Alternatively, imagine (perhaps implausibly) that the research shows no evidence that the differences between positive labels for men and women in the historic data are the result of unjust social structures. In other words, the bank concludes that any differences are the result of factors for which we would hold the individuals personally responsible. In this case, the bank could apply an individual fairness measure which does not attempt to adjust individual features on the basis of gender. A man and woman with the same repayment record would be regarded as similar, because there is no evidence of (for instance) gender-unequal leniency of bank clerks.

But given these same assumptions, they could also apply group-based fairness measures like equal calibration. As with the college admissions example, equal calibration does not treat disparities in outcomes (in this case, loan repayment) as evidence of unfairness which the model should avoid learning and reproducing (instead, it only addresses disparities in calibration between genders). As such, it is unlikely that an equally-calibrated model would (other things being equal) give a more positive prediction to a woman than a man if they both have equal repayment rates. As such, equal calibration is a group-fair measure which preserves the same kind of normative considerations that an un-adjusted individual fairness would aim to protect in this scenario (and ignores the normative considerations that motivate other group-fair measures like statistical parity and individual-fair measures that incorporate adjustments to features based on protected group membership).

\subsection{The important difference is between worldviews, not between individual or group fairness measures}

We do not deny that there are important differences between these measures in their implementation. In particular, the lack of any clear process for how decision-makers should determine an appropriate individual similarity metric makes individual fairness a particularly difficult measure to implement in practice. Conversely, statistical parity is easy, and straightforward to justify where decision-makers simply use differing base rates between groups as an indirect proxy for the level of structural discrimination affecting them. But despite these differences, once their starting assumptions are clear, decision-makers will be able to find variants within both the individual and group fairness families to choose from that reflect those assumptions (see Table 2).

A consequence of this view is that the supposed conflict between two broad families of fairness measure is actually only a conflict between \emph{specific variants} within them, which represent two different worldviews. The `raw' similarity metric reflects the worldview that disparities are due to personal choices. This particular variant of individual fairness conflicts with the statistical parity variant of group fairness, which reflects the view that disparities are due to unjust structures.

By selecting a different pair of variants from each family, we can reverse the polarity. A \emph{group-adjusted} variant of individual fairness, which adjusts individuals' features based on their group membership (e.g. adding extra SAT points to members of a disadvantaged group), reflects the view that disparities are due to unjust structures. This would conflict with the \emph{equal calibration} variant of group fairness, which reflects the view that disparities in base rates in the data are due to personal choices (or some other morally benign cause). Individual and group fairness measures can therefore appear on either side of the conflict depending on which variants we select. The same conflict can of course also arise between measures within the same family. For instance, in the infamous example of the COMPAS criminal recidivism risk scoring algorithm, it was illustrated that it is impossible to simultaneously achieve two group-fairness measures; namely, false positive equality (which reflects the assumption of structural bias) and equal calibration (which reflects the assumption that disparities are benign) \cite{chouldechova2017fair}.

\subsubsection{On the legitimacy of decision-makers normative assumptions}

It is of course both empirically and politically difficult for decision-makers to establish agreed assumptions about which factors are the result of an individual's free choice or (un)lucky circumstance, and the extent to which these factors account for differences in base rates of observed features between groups. A related and potentially even more politically fraught question concerns the extent to which decision-makers in particular contexts should be held responsible for compounding or actively correcting those inequalities. 

Aside from these problems, the very framing of Fair-ML can also be criticised on the grounds that it centres the decision-maker and assumes the legitimacy of their power to make decisions based on algorithms, including choosing which contestable assumptions to incorporate into them \cite{hoffmann2019fairness,powles2018}. In many cases such legitimacy is rightly challenged. Even if the decision-maker makes the `right' assumptions about the extent of discrimination in a particular context and honestly and competently factors that into their model, we can still question whether they should have the power to wield an ML model to make consequential decisions in the first place. Indeed, confronted with the necessity to make normative and societal assumptions which go beyond the data, they may conclude that the only viable option is to reject the use of an ML system altogether.

However, if there are at least \emph{some} cases in which ML systems are legitimate, then in order to have meaningful and justifiable fairness constraints at all, Fair-ML practitioners cannot avoid making some contestable empirical and normative assumptions.

\subsubsection{Relation to the (Im)Possibility of Fairness}

The point above - that it is necessary to make contestable empirical and normative assumptions in order to pursue the aims of Fair-ML in a meaningful way - is closely related to arguments put forward in previous work by Friedler et al. \cite{friedler2016possibility}. The authors introduce a useful set of distinctions between possible worldviews that could be assumed (also using the hypothetical example of college admissions decisions). Admissions tutors might decide they want to base their decisions on predicted success of applicants, and posit that attributes like self-control and `grit' are determining factors of success. Such attributes are represented in `construct space' - the space of features that the decision-maker would like to use to make a decision. Ideally, the decision-maker would prefer to be able to map from features in construct space directly to decision space (e.g. admit an applicant to college or not). In practice, they must rely on an `observed space' containing imperfect but measurable proxies for the features in construct space (e.g. survey-based self-control scores).

The authors propose two possible world-views with significantly different consequences for how we approach Fair-ML. On one view, we could assume that the observed space accurately maps onto the construct space; in other words, \emph{`What You See is What You Get'} (WYSIWYG). On another view, \emph{`We're All Equal'} (WAE), any differences in distributions of features between groups we assume are due to structural bias, which means that the distances between individuals from different groups may become distorted during the transformation between construct and observed space (a phenomenon they term `group skew').\footnote{Other worldviews not mentioned in \cite{friedler2016possibility} are also imaginable; for instance, rather than assuming WYSIWYG or WAE, one might take the view that even though a certain group is already disproportionately likely to have positive outcomes in observed space compared to other groups, they may actually be \emph{under}-represented given their `real' representation in construct space. Or it might be that once structural bias is fully accounted for, members of usually under-represented groups might be \emph{even more} qualified than the general population for the task at hand (`What You See is the Oppposite of What You Get').}

These are equivalent to the contrasting assumptions introduced in the examples above. Friedler et. al. argue that we \emph{have} to make assumptions about the extent to which each of these worldviews are true, in order to meaningfully implement fairness constraints. Although they do not address the relationship between different worldviews and the individual / group fairness distinction, focusing on assumed worldviews in this way can help illustrate why apparent conflicts between individual and group fairness are misguided. The important difference is between worldviews, not whether we render our assumptions about them at the individual or group level. The reason for treating people differently on the basis of group membership is the assumption that it is comparatively harder for members of group A to get a certain score - i.e. there is group skew when transforming between construct and observed space. Any deviations from (or adjustments to) a task-relevant similarity metric have therefore already been justified. Unsuccessful applicants from group B cannot complain that they have been treated unfairly because they have better `raw' scores than successful applicants from group A, unless they actually intend to disagree with the WAE assumption. 


Similarly, if we assume there is \emph{no} structural discrimination (WYSIWYG), then the observed space can be assumed to transform evenly onto construct space. In which case, there are both group and individual measures which could be appropriate. We could choose a similarity metric where distance in feature space is taken as an unproblematic correspondence to task-relevant feature distance. Or alternatively, we could choose a group fairness measure like equal calibration, which allows for differences in outcomes between groups in so far as the underlying base rates in labels differ. Either way, on this view, observed space is assumed to very closely correspond to construct space.





\section{Conclusion}

To summarise the arguments above: despite their apparent conflict, individual and group fairness measures do not necessarily reflect different normative principles. Both are commensurate with notions of consistency and egalitarianism derived from legal and political philosophy. Both fail to satisfy the principle of individual justice (despite that principle's surface-level similarities with individual fairness).

In practice, attempts to implement individual or group fairness would have to go through the same set of questions. These include questions about the purpose of the algorithmic system (e.g. `what kinds of candidates do we want the system to select?'), about the underlying assumptions regarding the data-generating process (`how are qualifications obtained and recorded?'), and about the kinds of unfairness at issue (`how have structural inequalities affected the distribution of features between the relevant sub-groups?').

The supposed conflict between individual and group fairness measures is more an artefact of the fact that different answers to these empirical and normative questions are typically associated with one or the other family of measure. But there are versions of individual and group fairness which can satisfy whatever assumptions we might have. This is not to imply that there are no differences between the two families of measure in terms of the way they are implemented. However, neither family can be exclusively associated with one set of ethical, social and causal assumptions. As a result, the normative differences \emph{within} the two families of Fair-ML measures are more important than the differences \emph{between} them.

As well as helping to dissolve apparent conflicts between individual and group fairness, thinking about the normative and empirical assumptions behind particular fairness measures may lead to other, more productive questions. For instance, if the assumption behind the application of statistical parity approaches is that the data reflect unjust structures, how might we change those underlying structures (rather than papering over them by tinkering with the model)? What could be done to intervene on the data-generating process that would also challenge structural oppression more directly? Finally, if the need to `treat people as individuals' really is the underlying motivation for adopting individual fairness measures, then perhaps we should reconsider whether to use ML in such contexts at all, if it is indeed incompatible with individual justice.

\section{Acknowledgements}
This work was supported by the UK Engineering and Physical Sciences Research Council (EPSRC) under grant number N02334X/1.

\bibliographystyle{ACM-Reference-Format}
\bibliography{sample-base}

\end{document}